\documentclass[11pt,a4paper]{article}
\usepackage{times,latexsym}
\usepackage{url}
\usepackage[T1]{fontenc}

\usepackage[acceptedWithA]{tacl2021v1}
\AddToShipoutPicture{\AtPageLowishCenter{\thepage}}

\usepackage{amsmath}
\usepackage{amssymb}
\usepackage{amsthm}        
\usepackage{booktabs}
\usepackage{algorithm}
\usepackage{algpseudocode}

\theoremstyle{definition}

\usepackage{xcolor}
\newcommand{\lpo}{PLPO}  

\usepackage{xspace,mfirstuc,tabulary}

\newif\iftaclinstructions
\taclinstructionsfalse 
\iftaclinstructions
\renewcommand{\confidential}{}
\renewcommand{\anonsubtext}{(No author info supplied here, for consistency with
TACL-submission anonymization requirements)}
\newcommand{\instr}
\fi

\iftaclpubformat 

\else

\fi

\title{Improving the Chain-of-Thought Compressibility of Pruned Reasoning Models
via Long--Short Contrastive Pruning}

\author{
  Tao Jing,
  Chen Kang,
  Feifei Zhai,
  Pengyuan Liu,
  Dong Yu
}

\date{}

\begin{document}
\maketitle

\begin{abstract}
  Large reasoning models are costly to serve because they are large and because they generate long chains of thought (CoT). Weight pruning and CoT compression address these two costs separately, but they interfere in that pruning adds redundant reasoning on problems the model still solves and incomplete reasoning on those it no longer solves, and on harder problems makes the model less compressible by subsequent length-preference optimization. We trace this to a blind spot on each side: pruning criteria do not distinguish necessary from redundant reasoning, and preference-based CoT compression receives no training signal on problems a pruned model always or never solves. We propose Contrastive Set-Difference (CSD), which identifies neurons enriched in redundant reasoning by contrasting their activations on long and short correct solutions of the same problem and prunes them preferentially, and we extend preference pairs to the problems a pruned model always or never solves. Pruning 20\% of DeepSeek-R1-Distill-Qwen-7B, the combination uses 8--22\% fewer thinking tokens on four math benchmarks than pruning with reasoning-aware calibration (RAC) followed by the same compression, with accuracy within 0.5 points.
\end{abstract}
 
\section{Introduction}
\label{sec:intro}
 
Large reasoning models (LRMs) solve hard problems by generating long chains of thought (CoT)~\citep{openai2024o1,guo2025deepseekr1}. Their inference cost therefore depends on two factors. One is the cost of each generated token, which grows with model size, and the other is the number of tokens generated. Two lines of work reduce these factors separately. Weight pruning removes parameters to cut memory and per-token compute~\citep{frantar2023sparsegpt,sun2024wanda,an2024flap}, and CoT compression shortens reasoning, typically by training the model to reach its answers in fewer tokens~\citep{kimi2025k15,luo2025o1pruner,aggarwal2025l1,hou2025thinkprune,zhang2025tokensqueeze}. A deployment that needs both a smaller model and shorter outputs would combine the two and, ideally, multiply their savings.
 
In practice, the two interfere. \citet{lucas2026rac} observed that pruning can make reasoning models generate more thinking tokens at lower accuracy, at times making them slower than the dense model. 
The added length has two sources. On problems the model still solves, it produces more redundant reasoning, that is, content a shorter correct solution of the same problem does without.
This is the overthinking that dense reasoning models already show on simple problems~\citep{chen2025overthinking}, amplified by pruning. 
On problems it no longer solves, it produces incomplete reasoning that runs long without reaching a correct answer. Pruning also makes the model harder to shorten: on harder problems, the same length-preference optimization reduces the thinking length of a pruned model less than that of the dense model. We call this relative reduction the model's compressibility.
 
We trace this interference to a blind spot on each side. Pruning methods for reasoning models, including RAC, RESP, and LayerTrim~\citep{lucas2026rac,wang2025resp,layertrim2026}, score units by their importance on the model's own reasoning traces.
Such scores do not distinguish necessary from redundant reasoning, so part of the budget goes to neurons whose activity is enriched in redundant reasoning. CoT-compression methods, in turn, take the model as given and rely on contrasting samples of it, which a pruned model often cannot provide.
No prior work asks how pruning shapes what can be compressed afterwards.
 
We address both blind spots with two drop-in components. On the pruning side, Contrastive Set-Difference (CSD) contrasts activations on long and short correct solutions of the same problem to identify neurons enriched in redundant reasoning, subtracts a protected set of high-importance neurons, and prunes the remainder without changing the pruned architecture or budget. On the compression side, we extend TokenSqueeze's preference pairs to the problems they miss: where the pruned model never succeeds, the dense model's shortest correct solution serves as the chosen response; where it never fails, its own longer correct solutions serve as rejected responses. Training uses TokenSqueeze's objective unchanged. Both components thus share one notion of redundancy: the excess over the shortest correct solution.
 
At 20\% sparsity on four math benchmarks, CSD with our preference pairs uses 8--22\% fewer thinking tokens than RAC-calibrated pruning followed by the same compression, with accuracy within 0.5 points. Before compression, CSD already shortens reasoning on MATH-500 and GSM8K, with accuracy within 0.4 points, and the same compression shortens it more than it shortens RAC-calibrated pruning on every benchmark.
 
Our analyses explain where the gains come from and where they stop. The neurons CSD removes form a distributed population rather than a few dedicated units, and they are not redundant as parameters: the base criterion is less likely to prune them than an average neuron. Before compression, the saved tokens come from shorter completed reasoning on problems the model still solves.
 
Our contributions are as follows:
\begin{itemize}
    \item We identify how weight pruning interferes with CoT compression: it adds redundant and incomplete reasoning, and it reduces the model's compressibility under length-preference optimization.
    \item We propose CSD, a contrastive set-difference criterion that identifies neurons enriched in redundant reasoning from long and short correct solutions of the same problem, together with a pruning-aware construction of preference pairs that covers problems the pruned model always or never solves.
\end{itemize}
 
\section{Related Work}
\label{sec:related}

\paragraph{Reasoning-aware pruning.}
Post-training pruning removes weights using importance scores estimated on a small calibration set. Unstructured and semi-structured methods sparsify individual weights~\citep{frantar2023sparsegpt,sun2024wanda}, while structured methods remove neurons, attention heads, or hidden dimensions to obtain direct speedups~\citep{ma2023llmpruner,ashkboos2024slicegpt,an2024flap} and are often followed by recovery training or distillation~\citep{xia2024sheared,muralidharan2024minitron}. The choice of calibration data strongly affects the outcome, especially at high sparsity~\citep{williams2024calibration,ji2025beware}. Reasoning models are particularly sensitive. \citet{lucas2026rac} found that pruning can make them generate more thinking tokens at lower accuracy. Aggressive quantization likewise lengthens reasoning~\citep{quantoverthink2026}. Subsequent work adapts structured pruning to reasoning models through self-generated calibration with decode-time gradient importance~\citep{wang2025resp}, and \citet{zhang2026reasoningcompression} locate weights important for reasoning with difference-of-means and attribution patching. These methods retain the units the model needs to reproduce its reasoning, but none separates the reasoning a correct solution needs from the redundant reasoning it can do without.
 
\paragraph{CoT compression.}
Reasoning models often overthink, spending many tokens on redundant solutions to simple problems~\citep{chen2025overthinking}. CoT compression shortens reasoning by fine-tuning on shortened chains~\citep{kang2025c3ot,xia2025tokenskip,ma2025cotvalve,munkhbat2025selftraining}, by reinforcement learning with length rewards or constraints~\citep{kimi2025k15,luo2025o1pruner,aggarwal2025l1,hou2025thinkprune}, or by preference optimization over the model's own samples~\citep{shen2025dast,hong2026lcpo,zhang2025tokensqueeze}. Training-free methods instead steer activations~\citep{chen2025seal,huang2025manifold} or monitor neurons for early exit~\citep{liu2026neat} at inference time. All of these methods take the model as given; preference-based ones also need contrasting samples of the same model on the same problem, which a pruned model often cannot provide.

\section{Method}
\label{sec:method}
 
Pruning removes redundancy. For LLMs, redundancy is usually understood at the level of parameters: a neuron is redundant if removing it barely changes the model's computation. LRMs add a second kind of redundancy. Their outputs contain reasoning that a shorter correct solution of the same problem does without. The neurons that carry this reasoning are not redundant as parameters: they are active on the model's own reasoning traces, so reconstruction-based criteria keep them. We therefore treat the redundancy of an LRM as twofold, redundant neurons and neurons enriched in redundant reasoning, and prune both under the same budget. We then shorten the pruned model's reasoning with preference pairs built around its failure modes.

\subsection{Revisiting Redundancy in LRM Pruning}
\label{sec:prelim}
 
We prune the intermediate neurons of each MLP and keep all attention heads. Neuron $k$ of layer $\ell$ consists of row $k$ of $W_{\mathrm{gate}}$ and $W_{\mathrm{up}}$ and column $k$ of $W_{\mathrm{down}}\in\mathbb{R}^{d\times I}$. Let $a_{t,k}$ be the $k$-th input to $W_{\mathrm{down}}$ at token $t$, and let $M_k(\mathcal{T})=\frac{1}{|\mathcal{T}|}\sum_{t\in\mathcal{T}}a_{t,k}^2$ be its mean square over a token set $\mathcal{T}$. Following the structured Wanda score of FLAP~\citep{an2024flap}, the base score of neuron $k$ is
\begin{equation}
s_k=\frac{1}{d}\sum_{i=1}^{d}\bigl|W_{\mathrm{down}}[i,k]\bigr|\cdot\sqrt{M_k(\mathcal{T})},
\label{eq:base}
\end{equation}
a proxy for the output change caused by removing the neuron. Following RAC~\citep{lucas2026rac}, $\mathcal{T}$ contains the prompts and the model's own chains of thought. Each layer removes the $B$ neurons with the smallest $s_k$. Layers are pruned sequentially, and calibration activations are propagated through the already pruned layers. Removed neurons are physically deleted, so the result is a smaller dense model.

Neurons with small $s_k$ are redundant neurons: the model barely uses them on its own reasoning traces. The second kind of redundancy lives in the output.

Let $y^{\mathrm{S}}$ and $y^{\mathrm{L}}$ be two correct solutions of a problem $x$ with $|y^{\mathrm{S}}|<|y^{\mathrm{L}}|$. We refer to the length excess of $y^{\mathrm{L}}$ over $y^{\mathrm{S}}$ as redundant reasoning: content that a correct solution of the same problem does without. Because the calibration traces contain redundant reasoning, the base criterion tends to keep neurons neurons that enriched in redundant reasoning. Structured pruning of LRMs should therefore consider both kinds of redundancy.
 
\subsection{CSD: Contrastive Set-Difference}
\label{sec:csd}
 
\paragraph{Contrastive objective.}
Let $\mathcal{D}_{\mathrm{S}}$ and $\mathcal{D}_{\mathrm{L}}$ be short and long correct solutions of the same problems, and let $M^{\mathrm{S}}_k$ and $M^{\mathrm{L}}_k$ be the mean squares on the two streams. Removing a set $R$ of neurons changes the MLP output on stream $X$ by
\begin{equation}
\begin{split}
E_X(R) &= \mathbb{E}_{t\in X}\Bigl\|\sum_{k\in R}a_{t,k}\,W_{\mathrm{down}}[:,k]\Bigr\|_2^2\\
&\approx \sum_{k\in R}\|W_{\mathrm{down}}[:,k]\|_2^2\,M^X_k,
\end{split}
\label{eq:err}
\end{equation}
where the approximation ignores cross-neuron terms. We seek to preserve the computation on short solutions while discounting the computation that long solutions add:
\begin{equation}
J_\lambda(R)=(1+\lambda)\,E_{\mathrm{S}}(R)-\lambda\,E_{\mathrm{L}}(R).
\label{eq:obj}
\end{equation}
The cost of removing neuron $k$ then becomes
\begin{equation}
c^{(\lambda)}_k=\|W_{\mathrm{down}}[:,k]\|_2^2\,M^{\mathrm{L}}_k\,g_\lambda(r_k),
\label{eq:glambda}
\end{equation}
where $g_\lambda(r)=(1+\lambda)/r-\lambda$ and $r_k=M^{\mathrm{L}}_k/M^{\mathrm{S}}_k$. The contrast enters only through this ratio. The family contains familiar choices: $\lambda=-1$ calibrates on long solutions only, which is RAC; $\lambda=-\beta\in(-1,0)$ mixes the two streams' mean squares with weight $\beta$; $\lambda=0$ calibrates on short solutions only. For $\lambda>0$, $g_\lambda$ decreases in $r$ and becomes negative once $r_k>(1+\lambda)/\lambda$, so removing such a neuron lowers the objective. To see what $r_k$ measures, write a long solution as a core $K$ that the short solution shares plus an excess $R$ that makes up a fraction $\rho$ of its tokens. If $M^{\mathrm{S}}_k\approx M^{K}_k$, then
\begin{equation}
r_k-1\approx\rho\left(\frac{M^{R}_k}{M^{K}_k}-1\right),
\label{eq:kr}
\end{equation}
so $r_k>1$ indicates that neuron $k$ is more active on redundant reasoning than on the shared core.
 
\paragraph{From a global $\lambda$ to a set difference.}
A single $\lambda>0$ for all neurons is undesirable: $r_k$ is a noisy ratio close to one for most neurons, calibration that leans on short solutions ($\lambda\ge-1/2$) does not improve on RAC, and neurons important to all reasoning, to concluding a solution, or to the model's numerics must not be displaced. We therefore set $\lambda$ per neuron: $\lambda_k\to+\infty$, that is, forced removal, for unprotected neurons with significant enrichment, and $\lambda_k=-1$ for all others. This is a set difference in the spirit of \citet{wei2024brittleness}.
 
\paragraph{Estimating the contrast.}
Among the solutions sampled from the dense model for each problem, $y^{\mathrm{S}}$ is the shortest correct one and $y^{\mathrm{L}}$ is drawn uniformly from the other correct, normally terminated ones; problems with fewer than two correct solutions are discarded. Pairing within a problem controls for difficulty. Statistics are accumulated over generated tokens only (reasoning, \texttt{</think>}, and answer), because prompts are identical within a pair. Long solutions occupy later positions, so we match positions: tokens are binned by absolute position in the full sequence, with bin edges at 512, 1024, 2048, 4096, and 8192. Let $\pi_b$ be the share of short-stream tokens in bin $b$, renormalized over the set $\mathcal{B}_\cap$ of bins that both streams cover, and let $\hat M^X_k=\sum_{b\in\mathcal{B}_\cap}\pi_b\,M^X_k(b)$ be the position-matched mean square for $X\in\{\mathrm{S},\mathrm{L}\}$. The log-ratio is
\begin{equation}
\hat\kappa_k=\log\bigl(\hat M^{\mathrm{L}}_k+\delta\bigr)-\log\bigl(\hat M^{\mathrm{S}}_k+\delta\bigr),
\label{eq:ratio}
\end{equation}
where $\delta=0.01\cdot\mathrm{median}_k\,M^{\mathrm{S}}_k$ is set per layer from the unmatched short-stream mean squares. For significance, we hash problems into $G=20$ groups, keeping each pair within one group, and use the $n$ groups with tokens in both streams. Recomputing Eq.~\eqref{eq:ratio} without group $g$, with $\pi_b$ re-estimated and $\delta$ fixed, gives $\hat\kappa^{(-g)}_k$, and the jackknife standard error is
\begin{equation}
\mathrm{SE}_k^2=\frac{n-1}{n}\sum_{g=1}^{n}\bigl(\hat\kappa^{(-g)}_k-\bar\kappa_k\bigr)^2,
\label{eq:z}
\end{equation}
where $\bar\kappa_k$ is the mean of the leave-one-out estimates; the test statistic is $z_k=\hat\kappa_k/\mathrm{SE}_k$.
 
\paragraph{Set-difference selection.}
In each layer, the protected set $P$ is the union of three sets:
\begin{itemize}
    \item the top 30\% of neurons by $s_k$;
    \item the top 1\% by closing ratio $e_k$, the mean square on the closing segment (from the last paragraph break before \texttt{</think>} to the end of the answer) divided by the mean square on all generated tokens, pooled over both streams;
    \item the top 0.1\% by massive activation $m_k=\max_t|a_{t,k}|$ over all calibration tokens, prompts included.
\end{itemize}
The candidate set and the displaced set are
\begin{equation}
\begin{aligned}
C &= \bigl\{k\notin P:\ z_k>z_0,\ \hat\kappa_k>\tau\bigr\},\\
D &= \mathrm{Top}_{m}\bigl(\{z_k\}_{k\in C}\bigr),
\end{aligned}
\label{eq:sets}
\end{equation}
where $m=\min\bigl(|C|,\lfloor qB\rfloor\bigr)$ and $\mathrm{Top}_m$ returns the $m$ neurons with the largest scores. The layer removes $D$ together with the $B-|D|$ neurons outside $D$ with the smallest $s_k$. The two parts of this set correspond to the two kinds of redundancy: $D$ holds neurons enriched in redundant reasoning, and the remainder holds redundant neurons. $P$ only restricts which neurons can be displaced; the base-score fill is applied as in RAC. When no neuron passes the test ($D=\emptyset$), CSD reduces exactly to RAC-calibrated structured pruning. CSD leaves the architecture and the per-layer budget unchanged. Its only extra cost is a teacher-forced pass over the short solutions, which are propagated through the pruned layers alongside the long ones.

\begin{algorithm}[t]
\caption{Contrastive Set-Difference (CSD) pruning}
\label{alg:csd}
\begin{algorithmic}[1]
\Require model with $L$ layers; paired streams $\mathcal{D}_{\mathrm{S}},\mathcal{D}_{\mathrm{L}}$; per-layer budget $B$; thresholds $z_0,\tau$; displacement ratio $q$
\For{$\ell=1,\dots,L$}
  \State Collect $a_{t,k}$ at the input of $W^{(\ell)}_{\mathrm{down}}$ on both streams (inputs come from the pruned layers $1,\dots,\ell-1$)
  \State Compute $s_k$ (Eq.~\eqref{eq:base}), $\hat\kappa_k$ (Eq.~\eqref{eq:ratio}), and $z_k$ (Eq.~\eqref{eq:z})
  \State $P\gets\mathrm{Top}_{30\%}(s)\cup\mathrm{Top}_{1\%}(e)\cup\mathrm{Top}_{0.1\%}(m)$
  \State $C\gets\{k\notin P: z_k>z_0,\ \hat\kappa_k>\tau\}$
  \State $D\gets\mathrm{Top}_{m}(\{z_k\}_{k\in C})$ with $m=\min(|C|,\lfloor qB\rfloor)$
  \State $R\gets D\ \cup$ the $B-|D|$ neurons outside $D$ with the smallest $s_k$
  \State Delete neurons $R$ from $W^{(\ell)}_{\mathrm{gate}},W^{(\ell)}_{\mathrm{up}},W^{(\ell)}_{\mathrm{down}}$ and propagate both streams through layer $\ell$
\EndFor
\end{algorithmic}
\end{algorithm}
 
\subsection{Pruning-Aware Length Preference Optimization}
\label{sec:pairs}
 
\paragraph{Sampling and outcomes.}
For each training problem $x$, we draw $N$ solutions from the pruned model $\tilde\pi$ and $N$ from the dense model $\pi$; the dense samples are shared across pruned models. A solution is \emph{correct} if its final answer is verified, it terminates normally, and it contains exactly one \texttt{</think>}. Solutions that are verified and terminate normally but do not contain exactly one \texttt{</think>} are discarded, and all others, including every truncated solution, are \emph{failures}. The length $|y|$ of a solution is its number of generated tokens, so a truncated solution has the length of the sampling budget. Let $\tilde{\mathcal{C}}(x)$ and $\tilde{\mathcal{F}}(x)$ be the pruned model's correct and failed solutions, and $\mathcal{C}(x)$ the dense model's correct ones.
 
\paragraph{Three sources of preference pairs.}
Every chosen response is a shortest correct solution, so all pairs share the notion of redundancy.
\begin{itemize}
    \item Self pairs follow TokenSqueeze~\citep{zhang2025tokensqueeze}. When $\tilde{\mathcal{C}}(x)$ and $\tilde{\mathcal{F}}(x)$ are both non-empty, the chosen responses are the $k$ shortest correct solutions, with $k=\lfloor0.2\,c(1-c/N)\rfloor+1$ and $c=|\tilde{\mathcal{C}}(x)|$, and the rejected responses are failures at least as long as the chosen one.
    \item Dense-anchored pairs cover problems that the pruned model never solves but the dense model does ($\tilde{\mathcal{C}}(x)=\emptyset\neq\mathcal{C}(x)$). The chosen response is the dense model's shortest correct solution, and the rejected responses are the pruned model's failures at least as long.
    \item Length-anchor pairs cover problems without self pairs on which the pruned model has at least two correct solutions; most of them it always solves. The chosen response is its shortest correct solution, and the rejected responses are its correct solutions at least $\gamma$ times longer.
\end{itemize}
 
\begin{table}[t]
\centering
\small
\begin{tabular}{@{}lll@{}}
\toprule
Pairs & Problems & Rejected $y_r$ \\
\midrule
Self   & $\tilde{\mathcal{C}}\neq\emptyset,\ \tilde{\mathcal{F}}\neq\emptyset$ & $y\in\tilde{\mathcal{F}},\ |y|\ge|y_c|$ \\
Dense  & $\tilde{\mathcal{C}}=\emptyset\neq\mathcal{C}$ & $y\in\tilde{\mathcal{F}},\ |y|\ge|y_c|$ \\
Length & no self pair, $|\tilde{\mathcal{C}}|\ge2$ & $y\in\tilde{\mathcal{C}},\ |y|\ge\gamma|y_c|$ \\
\bottomrule
\end{tabular}
\caption{Preference pairs: self, dense-anchored (Dense), and length-anchor (Length). Length-anchor pairs use problems without self pairs on which the pruned model has at least two correct solutions. The chosen response $y_c$ is the pruned model's shortest correct solution ($k$ shortest for self pairs), except for dense-anchored pairs, where it is the dense model's shortest correct solution. Self pairs target both redundant and incomplete reasoning, dense-anchored pairs target incomplete reasoning, and length-anchor pairs target redundant reasoning.}
\label{tab:pairs}
\end{table}
 
\paragraph{Objective.}
With $\pi_{\mathrm{ref}}$ the frozen starting model,
\begin{equation}
\begin{gathered}
\Delta_\theta=\log\frac{\pi_\theta(y_c|x)}{\pi_{\mathrm{ref}}(y_c|x)}-\log\frac{\pi_\theta(y_r|x)}{\pi_{\mathrm{ref}}(y_r|x)},\\
\Delta_{\mathrm{len}}=\log(1+|y_c|)-\log(1+|y_r|),\\
\mathcal{L}_{\mathrm{DPO\text{-}L}}=-\log\sigma\bigl(\beta\,[\Delta_\theta+\alpha\,\Delta_{\mathrm{len}}]\bigr),\\
\mathcal{L}_{\mathrm{SFT}}=-\tfrac{1}{|y_c|}\log\pi_\theta(y_c|x),\\
\mathcal{L}=(1-\eta)\,\mathcal{L}_{\mathrm{DPO\text{-}L}}+\eta\,\mathcal{L}_{\mathrm{SFT}}.
\end{gathered}
\label{eq:loss}
\end{equation}
Compared with TokenSqueeze, we sample fewer solutions per problem, under which its adaptive depth always selects the single shortest correct solution, and we omit its step-wise linguistic refinement. We also count truncated solutions as failures and add the dense-anchored and length-anchor pairs.

\section{Experiments}
\label{sec:exp}
 
\subsection{Setup}
\label{sec:setup}
 
\paragraph{Model and pruning.}
We prune DeepSeek-R1-Distill-Qwen-7B~\citep{guo2025deepseekr1} by removing 20\% of its decoder linear parameters. The removed parameters all belong to MLP neurons: every layer loses 4338 of its 18944 neurons (22.9\%), and all attention heads are kept.
 
\paragraph{Pruning criteria.}
We compare two criteria that share the per-layer budget, the base score of Eq.~\eqref{eq:base}, and the calibration data, and differ only in the displaced set $D$.
\emph{RAC-S} sets $D=\emptyset$, that is, structured Wanda-sp with the calibration of RAC~\citep{lucas2026rac}.
\emph{CSD} selects $D$ with Algorithm~\ref{alg:csd}, using $z_0=3$, $\tau=\log 1.1$, and $q=0.2$.
 
\paragraph{Calibration data.}
We sample 16 solutions per problem from the dense model for the first 1024 problems of OpenR1-Math~\citep{openr1}, at temperature 0.6 and top-$p$ 0.95 with at most 8192 tokens. 
Of the 585 problems with at least two correct solutions, we take the first 211 after shuffling. They give 1.00M tokens in the long stream and 0.70M in the short stream, prompts included, matching the 1M calibration tokens of RAC. Both criteria compute the base score on the long stream only ($\lambda=-1$ in Eq.~\eqref{eq:glambda}).
 
\paragraph{Length compression.}
We train each pruned model with \lpo{} on 3983 problems: a random 4000 of the 14{,}027 problems in the training set of TokenSqueeze~\citep{zhang2025tokensqueeze}, minus 17 near-duplicates of MATH-500 problems (normalized exact match or word-set Jaccard similarity of at least 0.8 against MATH-500, AIME24, and AIME25). We sample $N=8$ solutions per problem at temperature 0.9 and top-$p$ 1.0 with at most 8192 tokens. The dense model is sampled once, and its samples provide the chosen responses of dense-anchored pairs for both pruned models. Training updates all parameters for one epoch in batches of 32 pairs (about 400 steps), with AdamW, a constant learning rate of $5\times10^{-6}$ without warmup, gradient clipping at 1.0, bf16, $\alpha=1$, $\beta=0.1$, $\eta=0.5$, and $\gamma=1.5$; the reference policy is the frozen starting model. Problems, hyperparameters, and schedule are identical across starting points; only the starting model, and hence the sampled pairs, differ. We also apply \lpo{} to the dense model, for which dense-anchored pairs are empty, and evaluate the public TokenSqueeze-7B checkpoint~\citep{zhang2025tokensqueeze}, which its authors trained from the same dense model with more samples per problem and step-wise refinement.
 
\paragraph{Evaluation.}
We evaluate on MATH-500~\citep{hendrycks2021math,lightman2024verify} (500 problems) and GSM8K~\citep{cobbe2021gsm8k} (1,319 problems), using 4 samples per problem with a 16{,}384-token budget; and on AIME24 and AIME25 (30 problems each), using 16 samples per problem with a 32{,}768-token budget. 
Sampling uses vLLM with temperature 0.6 and top-$p$ 0.95. Each prompt is the problem wrapped in the model's chat template, with no system prompt or instruction. We report accuracy (Acc, pass@1 averaged over samples) and the mean number of thinking tokens (Think), which counts the tokens before the first \texttt{</think>}; a sample truncated before \texttt{</think>} counts at the full budget.

\begin{table*}[t]
\centering
\small
\setlength{\tabcolsep}{4pt}
\begin{tabular}{@{}lccccccccc@{}}
\toprule
& & \multicolumn{2}{c}{MATH-500} & \multicolumn{2}{c}{GSM8K} & \multicolumn{2}{c}{AIME24} & \multicolumn{2}{c}{AIME25} \\
\cmidrule(lr){3-4}\cmidrule(lr){5-6}\cmidrule(lr){7-8}\cmidrule(lr){9-10}
Model & Params & Acc & Think & Acc & Think & Acc & Think & Acc & Think \\
\midrule
\multicolumn{10}{@{}l}{\textit{Dense}} \\
Dense            & 7.62B & 92.9 & 3274 & 87.6 & 386 & 55.8 & 12781 & 43.1 & 14318 \\
TokenSqueeze-7B  & 7.62B & 89.8 & 1757 & 85.2 & 128 & 54.0 & 9378  & 38.3 & 10205 \\
Dense\,+\,\lpo{} & 7.62B & 90.7 & 1810 & 84.6 & 141 & 54.2 & 8357  & 41.5 & 9142 \\
\midrule
\multicolumn{10}{@{}l}{\textit{Pruned, before compression}} \\
RAC-S & 6.31B & 89.8 & 3780 & 85.7 & 799 & 42.3 & 16497 & 28.8 & 17321 \\
CSD   & 6.31B & 89.4 & 3671 & 85.7 & 735 & 41.9 & 16337 & 29.8 & 17416 \\
\midrule
\multicolumn{10}{@{}l}{\textit{Pruned, after \lpo{}}} \\
RAC-S\,+\,\lpo{} & 6.31B & 90.1 & 2636 & 81.9 & 277 & 46.5 & 13082 & 32.7 & 14533 \\
CSD\,+\,\lpo{}   & 6.31B & 89.6 & 2343 & 81.5 & 217 & 46.5 & 12010 & 32.9 & 12949 \\
\bottomrule
\end{tabular}
\caption{Main results. Acc is pass@1 (\%) and Think the mean number of thinking tokens, with samples truncated before \texttt{</think>} counted at the budget. Pruned models remove 20\% of decoder linear parameters, all from MLP neurons. }
\label{tab:main}
\end{table*}
 
\subsection{Pruning Alone}
\label{sec:res-prune}
 
The middle block of Table~\ref{tab:main} and the left half of Table~\ref{tab:paired} report the pruned models before compression.
 
\paragraph{Pruning lengthens reasoning.}
Compared with the dense model, RAC-S loses 3.1 points of accuracy on MATH-500, 1.9 on GSM8K, and 14.0 on AIME ($z=-6.5$). Its mean thinking length grows by 15.5\% on MATH-500 and by 24.8\% on AIME, where the truncation rate also rises by 13.1 points, and it doubles on GSM8K, from 386 to 799 tokens. On GSM8K, the added length comes almost entirely from samples that finish: their mean thinking length grows from 380 to 751 tokens while accuracy changes little. The pruned model is both weaker and more verbose, and the added verbosity is largest where problems are easiest.
 
\paragraph{CSD shortens reasoning on the easier benchmarks.}
On MATH-500, CSD thinks 4.4\% less than RAC-S in geometric mean and 2.9\% less in mean thinking length (3671 against 3780 tokens), with an accuracy difference of $-0.4$ points. On GSM8K, it thinks 8.0\% less (735 against 799) at the same accuracy. On AIME, it does not shorten reasoning: mean thinking length changes by $-0.2\%$ and accuracy by $+0.3$ points, while the truncation rate rises from 21.4\% to 24.0\%. The shortening is thus largest on the easiest benchmark.

\begin{table*}[t]
\centering
\small
\setlength{\tabcolsep}{4pt}
\begin{tabular}{@{}lccr@{\;}lccr@{\;}l@{}}
\toprule
& \multicolumn{4}{c}{Before \lpo{}} & \multicolumn{4}{c}{After \lpo{}} \\
\cmidrule(lr){2-5}\cmidrule(lr){6-9}
Benchmark & $\Delta$Acc & $\Delta$Think & \multicolumn{2}{c}{$\Delta$Geo} & $\Delta$Acc & $\Delta$Think & \multicolumn{2}{c}{$\Delta$Geo} \\
\midrule
MATH-500  & $-0.4$ & $-2.9\%$ & $-4.4\%$ & ($-3.0\sigma$)  & $-0.5$ & $-11.1\%$ & $-14.5\%$ & ($-7.3\sigma$)  \\
GSM8K     & $0.0$  & $-8.0\%$ & $-7.7\%$ & ($-6.6\sigma$)  & $-0.4$ & $-21.7\%$ & $-14.3\%$ & ($-13.3\sigma$) \\
AIME24+25 & $+0.3$ & $-0.2\%$ & $-0.1\%$ & ($-0.05\sigma$) & $+0.1$ & $-9.6\%$  & $-11.0\%$ & ($-3.6\sigma$)  \\
\bottomrule
\end{tabular}
\caption{CSD relative to RAC-S before and after \lpo{}. $\Delta$Acc is the difference in pass@1 in points and $\Delta$Think the relative change in mean thinking length, both computed from Table~\ref{tab:main}. $\Delta$Geo is the geometric mean over problems of the ratio of per-problem mean thinking lengths, minus one, computed from the per-problem results; in parentheses, the mean per-problem log ratio divided by its standard error over problems, which covers evaluation sampling only, not variation between training runs. AIME pools the two years with equal weight per problem.}
\label{tab:paired}
\end{table*}

\subsection{Compressing Pruned Models}
\label{sec:res-joint}

\paragraph{Pruning reduces compressibility on harder problems.}
On MATH-500 and AIME, the same recipe shortens the pruned model less than the dense model. \lpo{} reduces the mean thinking length of RAC-S by 30.3\% on MATH-500 and 18.3\% on AIME, against 44.7\% and 35.4\% for the dense model; on MATH-500, geometric means give the same ordering ($-42.9\%$ against $-63.9\%$). GSM8K is the exception: \lpo{} shortens RAC-S by 65.3\%, slightly more than the dense model (63.5\%), at a cost of 3.8 points of accuracy against 3.0 for the dense model. There, pruning adds mostly redundant reasoning on problems the model still solves, which the recipe removes; even so, RAC-S+\lpo{} still thinks about twice as long as Dense+\lpo{} (277 against 141 tokens). Pruning thus affects length in two ways: the pruned model starts longer and, where it adds incomplete reasoning, is harder to shorten.
 
\paragraph{CSD is more compressible than RAC-S.}
\lpo{} shortens CSD more than RAC-S on every benchmark. On MATH-500, the geometric-mean reduction is 48.9\% against 42.9\%; on the log scale, this recovers 24\% of the compressibility that RAC-S loses relative to the dense model. As a result, CSD+\lpo{} thinks 11.1\%, 21.7\%, and 9.6\% less than RAC-S+\lpo{} on MATH-500, GSM8K, and AIME, with accuracy differences of $-0.5$, $-0.4$, and $+0.1$ points (Table~\ref{tab:paired}). On MATH-500, the paired gap is 14.5\% in geometric mean.
 
\paragraph{Efficiency.}
Per generated token, the pruned models need 0.815$\times$ the linear-layer FLOPs of the dense model, including the output projection, and read 0.815$\times$ the weights. Attention is unchanged because all heads are kept, so the KV cache is the same for all models and the total cost ratio per token grows with context, from 0.82$\times$ at 1k tokens to 0.87$\times$ at 16k. The pruned models have 6.31B parameters against 7.62B (12.6 against 15.2\,GB in bf16). Combining per-token cost with thinking length, CSD+\lpo{} needs about 0.59$\times$ the dense decoding compute on MATH-500, against 0.55$\times$ for Dense+\lpo{} and 0.54$\times$ for TokenSqueeze-7B. The pruned models thus trade a slightly higher decoding cost than compression alone for 17\% fewer parameters.
 
\section{Analysis}
\label{sec:analysis}
 
We examine which neurons CSD removes, where the saved tokens come from, and whether the pruning side or the compression side matters more. All analyses use the 20\% sparsity models unless stated otherwise.

\subsection{What Does CSD Remove?}
\label{sec:ana-neurons}
 
\paragraph{CSD displaces neurons that RAC-S keeps.}
Across the 28 layers, CSD displaces 7719 neurons. Of these, 6456 lie above the RAC-S threshold and would have been kept; the other 1263 would have been removed anyway and leave the selection unchanged. CSD therefore changes 6456 of the 121{,}464 pruned neurons (5.3\%). The displaced neurons have a median base-score percentile of 46 (quartiles 30 and 59), while the neurons they replace sit just below the threshold, between the 20th and 23rd percentiles. The most enriched of them are spread over depth: of the top 10\% by $z$, a third lie in layers 18--23 and 14\% in layer 10 alone. These neurons rank in the middle of the base criterion because their activity on redundant reasoning counts as importance in Eq.~\eqref{eq:base}.

\paragraph{Their enrichment is spread over long solutions.}
We audit the 224 displaced neurons with the largest $z$, eight per layer, and split each neuron's enrichment over token types. For 53\% of them, no token contributes more than a quarter of the enrichment; 24\% have a reflection marker such as \emph{Wait} among their five largest contributors; and only 3\% owe at least half of it to a single rare token.
Most of the enrichment comes from stronger activation on the same tokens rather than from those tokens occurring more often in long solutions: this part accounts for a median 80\% of a neuron's enrichment.
The displaced neurons are therefore not detectors of specific words; they are more active throughout long solutions.

\subsection{Where Do the Saved Tokens Come From?}
\label{sec:ana-tokens}
 

We split the MATH-500 problems into three strata by mean thinking length in separate 8k-budget runs of the dense model and four lightly perturbed variants: the shorter half, the 50th to 80th percentiles, and the longest 20\%. The dense model solves 98.6\%, 97.8\%, and 71.0\% of the samples in these strata. 
Changes in length and accuracy can then be traced to two routes: redundant reasoning on problems the model still solves, and incomplete reasoning on problems it fails or truncates. Within strata, we report length changes as geometric means of per-problem ratios.

\paragraph{Pruning costs most on long problems.}
Relative to the dense model, RAC-S lengthens reasoning in every stratum, by 19.8\%, 13.4\%, and 16.0\%. 
Its accuracy loss and the rise in truncation, however, concentrate on the longest 20\% (10.5 points and 11.0 points); the middle stratum loses 3.2 points and the shorter half none. 

\paragraph{CSD shortens the problems the pruned model still solves.}
Relative to RAC-S, CSD shortens the shorter half by 5.2\% ($-3.2\sigma$). In the middle stratum and the longest 20\%, the changes are smaller and not significant ($-3.4\%$ and $-3.6\%$), and no stratum shows a significant accuracy difference (the largest is $-1.8$ points, $-1.8\sigma$, in the middle stratum). 
CSD thus acts on the redundant route: where the pruned model solves a problem reliably, it removes part of the excess over the shortest solution.

\subsection{Pruning-Side or Training-Side Optimization}
\label{sec:ana-side}
 
Recent work recovers structurally pruned reasoning models by on-policy distillation from the dense model. ReCal~\citep{jiang2026recal} removes FFN neurons amounting to 25\% of the parameters of Qwen3-8B with structured Wanda-sp and FLAP calibrated on prompts. Wanda-sp drops the mean AIME accuracy from 61.4\% to 2.2\%, and recovery reaches only 41.7\% for Wanda-sp and 29.0\% for FLAP. Recalibrating the same criterion on the dense model's responses, weighted by the damage pruning causes, raises the pruned model to 25.7\% and the recovered one to 48.5\%.
 
Recovery thus closes only part of the gap that pruning opens, and a better start ends higher. Calibrating on the model's own reasoning traces keeps most of the accuracy before any training. The models and protocols differ from those above, but the contrast is large: RAC-S stays within 3.1 points of the dense model on MATH-500 and 1.9 points on GSM8K, and keeps 35.5\% on AIME against 49.5\% (Table~\ref{tab:main}). Improving the pruning side therefore matters more than improving recovery.

\section{Conclusion}
\label{sec:conclusion}
 
Pruning and CoT compression target the two costs of serving a reasoning model, but pruning makes compression harder: it adds redundant reasoning on problems the model still solves and incomplete reasoning on those it no longer solves, and on harder problems it leaves the model less compressible. We proposed CSD, which prunes neurons enriched in redundant reasoning, identified by contrasting long and short correct solutions of the same problem, together with preference pairs that cover problems a pruned model always or never solves. At 20\% sparsity, the combination uses 8--22\% fewer thinking tokens than RAC-calibrated pruning followed by the same compression, with accuracy within 0.5 points, mostly because the model pruned by CSD is more compressible.
 
Two lessons follow. How a reasoning model is pruned matters more than how it is recovered, and pruning-side control of reasoning length works only while the pruned model keeps most of its capability. Our study covers one model and one training run per configuration. The pruned models also decode at a higher cost than the dense model compressed by the same recipe, in exchange for 17\% fewer parameters, so the combination pays off when memory rather than decoding compute is the binding constraint.

\bibliography{tacl2021} 

@article{openai2024o1,
  title={Openai o1 system card},
  author={Jaech, Aaron and Kalai, Adam and Lerer, Adam and Richardson, Adam and El-Kishky, Ahmed and Low, Aiden and Helyar, Alec and Madry, Aleksander and Beutel, Alex and Carney, Alex and others},
  journal={arXiv preprint arXiv:2412.16720},
  year={2024}
}

@article{guo2025deepseekr1,
  title={Deepseek-r1: Incentivizing reasoning capability in llms via reinforcement learning},
  author={Guo, Daya and Yang, Dejian and Zhang, Haowei and Song, Junxiao and Wang, Peiyi and Zhu, Qihao and Xu, Runxin and Zhang, Ruoyu and Ma, Shirong and Bi, Xiao and others},
  journal={arXiv preprint arXiv:2501.12948},
  year={2025}
}

@inproceedings{frantar2023sparsegpt,
  title={Sparsegpt: Massive language models can be accurately pruned in one-shot},
  author={Frantar, Elias and Alistarh, Dan},
  booktitle={International conference on machine learning},
  pages={10323--10337},
  year={2023},
  organization={PMLR}
}

@inproceedings{sun2024wanda,
  title={A simple and effective pruning approach for large language models},
  author={Sun, Mingjie and Liu, Zhuang and Bair, Anna and Kolter, Zico},
  booktitle={International Conference on Learning Representations},
  volume={2024},
  pages={4942--4964},
  year={2024}
}

@inproceedings{an2024flap,
  title={Fluctuation-based adaptive structured pruning for large language models},
  author={An, Yongqi and Zhao, Xu and Yu, Tao and Tang, Ming and Wang, Jinqiao},
  booktitle={Proceedings of the AAAI conference on artificial intelligence},
  volume={38},
  number={10},
  pages={10865--10873},
  year={2024}
}

@article{kimi2025k15,
  title={Kimi k1. 5: Scaling reinforcement learning with llms},
  author={Team, Kimi and Du, Angang and Gao, Bofei and Xing, Bowei and Jiang, Changjiu and Chen, Cheng and Li, Cheng and Xiao, Chenjun and Du, Chenzhuang and Liao, Chonghua and others},
  journal={arXiv preprint arXiv:2501.12599},
  year={2025}
}

@inproceedings{luo2025o1pruner,
  title={O1-pruner: Length-harmonizing fine-tuning for o1-like reasoning pruning},
  author={Luo, Haotian and He, Haiying and Wang, Yibo and Liu, Shiwei and Li, Wei and Cao, Xiaochun and Tao, Dacheng and Tan, Naiqiang and Shen, Li},
  booktitle={Findings of the Association for Computational Linguistics: ACL 2026},
  pages={14242--14257},
  year={2026}
}

@article{aggarwal2025l1,
  title={L1: Controlling how long a reasoning model thinks with reinforcement learning},
  author={Aggarwal, Pranjal and Welleck, Sean},
  journal={arXiv preprint arXiv:2503.04697},
  year={2025}
}

@article{hou2025thinkprune,
  title={Thinkprune: Pruning long chain-of-thought of llms via reinforcement learning},
  author={Hou, Bairu and Zhang, Yang and Ji, Jiabao and Liu, Yujian and Qian, Kaizhi and Andreas, Jacob and Chang, Shiyu},
  journal={arXiv preprint arXiv:2504.01296},
  year={2025}
}

@article{zhang2025tokensqueeze,
  title={TokenSqueeze: Performance-Preserving Compression for Reasoning LLMs},
  author={Zhang, Yuxiang and Yu, Zhengxu and Pan, Weihang and Jin, Zhongming and Fu, Qiang and Cai, Deng and Lin, Binbin and Ye, Jieping},
  journal={Advances in Neural Information Processing Systems},
  volume={38},
  pages={91516--91539},
  year={2026}
}

@inproceedings{lucas2026rac,
  title={Reasoning models can be accurately pruned via chain-of-thought reconstruction},
  author={Lucas, Ryan and Behdin, Kayhan and Wang, Zhipeng and Song, Qingquan and Mazumder, Rahul and others},
  booktitle={International Conference on Learning Representations},
  volume={2026},
  pages={142840--142852},
  year={2026}
}

@inproceedings{chen2025overthinking,
  title={Do NOT think that much for 2+ 3=? on the overthinking of long reasoning models},
  author={Chen, Xingyu and Xu, Jiahao and Liang, Tian and He, Zhiwei and Pang, Jianhui and Yu, Dian and Song, Linfeng and Liu, Qiuzhi and Zhou, Mengfei and Zhang, Zhuosheng and others},
  booktitle={Forty-second International Conference on Machine Learning},
  year={2025}
}

@article{wang2025resp,
  title={Think Before You Prune: Self-Reflective Structured Pruning for Reasoning Language Models},
  author={Wang, Ziyan and Diao, Enmao and Le, Qi and Wang, Pu and Wang, Guanchu and Lee, Minwoo and Yeh, Shu-ping and Yang, Li},
  journal={arXiv preprint arXiv:2512.02185},
  year={2025}
}

@article{layertrim2026,
  title={Unlocking Structured Pruning for Reasoning Language Models via Attention-Preserving MLP Trimming},
  author={Lim, Woosang and Yoo, Haneul and Yun, Vincent-Daniel and Yoo, Sungjoo},
  journal={arXiv preprint arXiv:2610.19121},
  year={2026}
}

@article{ma2023llmpruner,
  title={Llm-pruner: On the structural pruning of large language models},
  author={Ma, Xinyin and Fang, Gongfan and Wang, Xinchao},
  journal={Advances in neural information processing systems},
  volume={36},
  pages={21702--21720},
  year={2023}
}

@inproceedings{ashkboos2024slicegpt,
  title={Slicegpt: Compress large language models by deleting rows and columns},
  author={Ashkboos, Saleh and Croci, Maximilian and Gennari do Nascimento, Marcelo and Hoefler, Torsten and Hensman, James},
  booktitle={International Conference on Learning Representations},
  volume={2024},
  pages={11682--11701},
  year={2024}
}

@inproceedings{xia2024sheared,
  title={Sheared llama: Accelerating language model pre-training via structured pruning},
  author={Xia, Mengzhou and Gao, Tianyu and Zeng, Zhiyuan and Chen, Danqi},
  booktitle={International Conference on Learning Representations},
  volume={2024},
  pages={5385--5409},
  year={2024}
}

@article{muralidharan2024minitron,
  title={Compact language models via pruning and knowledge distillation},
  author={Muralidharan, Saurav and Turuvekere Sreenivas, Sharath and Joshi, Raviraj and Chochowski, Marcin and Patwary, Mostofa and Shoeybi, Mohammad and Catanzaro, Bryan and Kautz, Jan and Molchanov, Pavlo},
  journal={Advances in Neural Information Processing Systems},
  volume={37},
  pages={41076--41102},
  year={2024}
}

@inproceedings{williams2024calibration,
  title={On the impact of calibration data in post-training quantization and pruning},
  author={Williams, Miles and Aletras, Nikolaos},
  booktitle={Proceedings of the 62nd Annual Meeting of the Association for Computational Linguistics (Volume 1: Long Papers)},
  pages={10100--10118},
  year={2024}
}

@inproceedings{ji2025beware,
  title={Beware of calibration data for pruning large language models},
  author={Ji, Yixin and Xiang, Yang and Li, Juntao and Xia, Qingrong and Li, Ping and Duan, Xinyu and Wang, Zhefeng and Zhang, Min},
  booktitle={International Conference on Learning Representations},
  volume={2025},
  pages={18604--18620},
  year={2025}
}

@article{quantoverthink2026,
  title={Quantized reasoning models think they need to think longer, but they do not},
  author={Lotfi, Sanae and Kirichenko, Polina and Li, Steven and Liu, Zechun},
  journal={arXiv preprint arXiv:2606.00206},
  year={2026}
}

@inproceedings{zhang2026reasoningcompression,
  title={When reasoning meets compression: Understanding the effects of llms compression on large reasoning models},
  author={Zhang, Nan and Kwek, Eugene and Zhang, Yusen and Nguyen, Hieu and Mitra, Prasenjit and Zhang, Rui},
  booktitle={International Conference on Learning Representations},
  volume={2026},
  pages={62680--62706},
  year={2026}
}

@article{jiang2026recal,
  title={ReCal: Calibrating Structured Pruning for On-Policy Distillation Recovery},
  author={Jiang, Houcheng and Zheng, Mao and Song, Mingyang and Zhong, Qiyong and Sun, Jie and Zhang, Tianyu and Fang, Junfeng},
  journal={arXiv preprint arXiv:2610.11332},
  year={2026}
}

@inproceedings{kang2025c3ot,
  title={C3ot: Generating shorter chain-of-thought without compromising effectiveness},
  author={Kang, Yu and Sun, Xianghui and Chen, Liangyu and Zou, Wei},
  booktitle={Proceedings of the AAAI Conference on Artificial Intelligence},
  volume={39},
  number={23},
  pages={24312--24320},
  year={2025}
}

@inproceedings{xia2025tokenskip,
  title={Tokenskip: Controllable chain-of-thought compression in llms},
  author={Xia, Heming and Leong, Chak Tou and Wang, Wenjie and Li, Yongqi and Li, Wenjie},
  booktitle={Proceedings of the 2025 Conference on Empirical Methods in Natural Language Processing},
  pages={3351--3363},
  year={2025}
}

@inproceedings{ma2025cotvalve,
  title={Cot-valve: Length-compressible chain-of-thought tuning},
  author={Ma, Xinyin and Wan, Guangnian and Yu, Runpeng and Fang, Gongfan and Wang, Xinchao},
  booktitle={Proceedings of the 63rd Annual Meeting of the Association for Computational Linguistics (Volume 1: Long Papers)},
  pages={6025--6035},
  year={2025}
}

@inproceedings{munkhbat2025selftraining,
  title={Self-training elicits concise reasoning in large language models},
  author={Munkhbat, Tergel and Ho, Namgyu and Kim, Seo Hyun and Yang, Yongjin and Kim, Yujin and Yun, Se-Young},
  booktitle={Findings of the Association for Computational Linguistics: ACL 2025},
  pages={25127--25152},
  year={2025}
}

@inproceedings{shen2025dast,
  title={Dast: Difficulty-adaptive slow-thinking for large reasoning models},
  author={Shen, Yi and Zhang, Jian and Huang, Jieyun and Shi, Shuming and Zhang, Wenjing and Yan, Jiangze and Wang, Ning and Wang, Kai and Liu, Zhaoxiang and Lian, Shiguo},
  booktitle={Proceedings of the 2025 Conference on Empirical Methods in Natural Language Processing: Industry Track},
  pages={2322--2331},
  year={2025}
}

@inproceedings{hong2026lcpo,
  title={Pruning long chain-of-thought of large reasoning models via small-scale preference optimization},
  author={Hong, Bin and Liu, Jiayu and Zhang, Kai and Sun, Jianwen and Zhang, Mengdi and Huang, Zhenya},
  booktitle={International Conference on Learning Representations},
  volume={2026},
  pages={116818--116843},
  year={2026}
}

@article{chen2025seal,
  title={Seal: Steerable reasoning calibration of large language models for free},
  author={Chen, Runjin and Zhang, Zhenyu and Hong, Junyuan and Kundu, Souvik and Wang, Zhangyang},
  journal={arXiv preprint arXiv:2504.07986},
  year={2025}
}

@article{huang2025manifold,
  title={Mitigating overthinking in large reasoning models via manifold steering},
  author={Huang, Yao and Chen, Huanran and Ruan, Shouwei and Zhang, Yichi and Wei, Xingxing and Dong, Yinpeng},
  journal={Advances in Neural Information Processing Systems},
  volume={38},
  pages={102543--102568},
  year={2026}
}

@inproceedings{liu2026neat,
  title={Neat: Neuron-based early exit for large reasoning models},
  author={Liu, Kang and Liu, Yongkang and Yang, Xiaocui and Wang, Peidong and Zhang, Wen and Feng, Shi and Zhang, Yifei and Wang, Daling},
  booktitle={Findings of the Association for Computational Linguistics: ACL 2026},
  pages={24616--24627},
  year={2026}
}

@article{wei2024brittleness,
  title={Assessing the brittleness of safety alignment via pruning and low-rank modifications},
  author={Wei, Boyi and Huang, Kaixuan and Huang, Yangsibo and Xie, Tinghao and Qi, Xiangyu and Xia, Mengzhou and Mittal, Prateek and Wang, Mengdi and Henderson, Peter},
  journal={arXiv preprint arXiv:2402.05162},
  year={2024}
}

@article{hendrycks2021math,
  title={Measuring mathematical problem solving with the math dataset},
  author={Hendrycks, Dan and Burns, Collin and Kadavath, Saurav and Arora, Akul and Basart, Steven and Tang, Eric and Song, Dawn and Steinhardt, Jacob},
  journal={arXiv preprint arXiv:2103.03874},
  year={2021}
}

@inproceedings{lightman2024verify,
  title={Let's verify step by step},
  author={Lightman, Hunter and Kosaraju, Vineet and Burda, Yuri and Edwards, Harrison and Baker, Bowen and Lee, Teddy and Leike, Jan and Schulman, John and Sutskever, Ilya and Cobbe, Karl},
  booktitle={International Conference on Learning Representations},
  volume={2024},
  pages={39578--39601},
  year={2024}
}

@article{cobbe2021gsm8k,
  title={Training verifiers to solve math word problems},
  author={Cobbe, Karl and Kosaraju, Vineet and Bavarian, Mohammad and Chen, Mark and Jun, Heewoo and Kaiser, Lukasz and Plappert, Matthias and Tworek, Jerry and Hilton, Jacob and Nakano, Reiichiro and others},
  journal={arXiv preprint arXiv:2110.14168},
  year={2021}
}

@misc{openr1,
  title  = {Open {R1}: A fully open reproduction of {DeepSeek-R1}},
  url    = {https://github.com/huggingface/open-r1},
  author = {{Hugging Face}},
  month  = {January},
  year   = {2025}
}
\bibliographystyle{acl_natbib}

\end{document}